\documentclass[a4paper,conference]{IEEEtran}

\usepackage{cite}
\usepackage{booktabs} 
\ifCLASSINFOpdf
  \usepackage[pdftex]{graphicx}
\else
  \usepackage[dvips]{graphicx}
\fi
\usepackage{amsmath}
\usepackage{subfig}
\usepackage{algorithm}
\usepackage{algorithmic}
\usepackage{array}
\usepackage{fixltx2e}
\usepackage{stfloats}
\usepackage{url}

\begin{document}

\title{Adversarially Robust Classification by Conditional Generative Model Inversion}

\author{\IEEEauthorblockN{Mitra Alirezaei}
\IEEEauthorblockA{Department of Electrical and\\Computer Engineering\\
University of Utah\\
Salt Lake City, UT, USA \\
Email: mitra@sci.utah.edu}
\and
\IEEEauthorblockN{Tolga Tasdizen}
\IEEEauthorblockA{Department of Electrical and\\Computer Engineering\\
University of Utah\\
Salt Lake City, UT, USA \\
Email: tolga@sci.utah.edu}}

\maketitle

\begin{abstract}
Most adversarial attack defense methods rely on obfuscating gradients. These methods are successful in defending against gradient-based attacks; however, they are easily circumvented by attacks which either do not use the gradient or by attacks which approximate and use the corrected gradient. Defenses that do not obfuscate gradients such as adversarial training exist, but these approaches generally make assumptions about the attack such as its magnitude. We propose a classification model that does not obfuscate gradients and is robust by construction without assuming prior knowledge about the attack. Our method casts classification as an optimization problem where we "invert" a conditional generator trained on unperturbed, natural images to find the class that generates the closest sample to the query image. We hypothesize that a potential source of brittleness against adversarial attacks is the high-to-low-dimensional nature of feed-forward classifiers which allows an adversary to find small perturbations in the input space that lead to large changes in the output space. On the other hand, a generative model is typically a low-to-high-dimensional mapping. Since the range of images that can be generated by the model for a given class is limited to its learned manifold, the "inversion" process cannot generate images that are arbitrarily close to adversarial examples leading to a robust model by construction. While the method is related to Defense-GAN, which cleans adversarial examples by projecting to the data manifold prior to passing them to a classifier, the use of a conditional generative model and inversion in our model instead of the feed-forward classifier is a critical difference. Unlike Defense-GAN, which was shown to generate obfuscated gradients that are easily circumvented, we show that our method does not obfuscate gradients. We demonstrate that our model is extremely robust against black-box attacks and has improved robustness against white-box attacks compared to naturally trained, feed-forward classifiers. 
\end{abstract}


%
\IEEEpeerreviewmaketitle

\section{Introduction}
In the last decade, deep learning has achieved unprecedented success in many fields. For instance, convolutional neural networks (CNN) have revolutionized the field of computer vision \cite{krizhevsky2012imagenet,DBLP:journals/corr/SimonyanZ14a,DBLP:journals/corr/SzegedyLJSRAEVR14,lecun2015deep,Ren2015,Redmon2016,he2016deep}. However, they are vulnerable against \emph{adversarial attacks} \cite{szegedy2013intriguing, goodfellow2014explaining}. Adversarial attacks are carefully crafted perturbations, which when added to a natural sample, fool the classifier, and at the same time, they do not affect human recognition. Adversarial examples pose a security concern in applications such as autonomous driving and healthcare and are a critical road block to wider adoption of these methods in many fields that could greatly benefit from their use.


Researchers have developed various methods to defend against adversarial attacks; however, a large subset of these methods rely on, intentionally or unintentionally, obfuscating gradients of the model. The notion of \emph{obfuscated gradients} was introduced in \cite{athalye2018obfuscated} as a special case of gradient masking \cite{papernot2017practical}. Obfuscated gradients occur when the defense method breaks the gradients or introduces nonexistent gradients. Consequently, iterative optimization attacks that rely on gradients are not successful. For instance, applying input transformations to the input \cite{guo2017countering} and
input prepossessing to remove adversarial perturbations from input samples \cite{samangouei2018defense, song2017pixeldefend} increase the robustness. However, these methods were shown to defend against adversarial attacks by obfuscating gradients \cite{athalye2018obfuscated}. Methods based on obfuscating gradients do not eliminate adversarial examples but rather confound them \cite{athalye2018obfuscated, xu2020adversarial}. Athalye et al.  \cite{athalye2018obfuscated} introduced techniques to circumvent defense techniques that rely on obfuscated gradients.  Backward pass differentiable approximation (BPDA) is one of these attack techniques that was introduced to overcome gradient shattering. BPDA approximates the gradient by performing the forward pass and approximating the backward pass using a differentiable approximation of the function. BPDA was shown to break down defense mechanisms with obfuscated gradients. Other methods which obfuscate gradients and were circumvented by Athalye et al.  \cite{athalye2018obfuscated}  include Defense-GAN, PixelDefend, the  \cite{guo2017countering} method, local intrinsic dimensionality (LID) \cite{ma2018characterizing}, stochastic activation pruning (SAP) \cite{dhillon2018stochastic}, mitigating through randomization \cite{xie2017mitigating}, and thermometer encoding \cite{buckman2018thermometer} . 

Defenses that do not obfuscate gradients such as adversarial training exist, but these approaches generally make assumptions about the attack such as its magnitude. In adversarial training, the training data is augmented with adversarial examples \cite{szegedy2013intriguing, goodfellow2014explaining}. Adversarial training is one of the most promising defenses; however, it is not effective against stronger attacks or non-gradient-based attacks.

We introduce a new approach to classify images without obfuscating gradients or assumptions about attack type or strength. We invert a conditional generator to classify images.

First, a conditional generative adversarial network (cGAN) \cite{mirza2014conditional} is trained to model the distribution of unperturbed images. The trained conditional generator is then used to generate images similar to the query image to be classified by inverting the generator \cite{creswell2018inverting}. Inverting the generator is performed by finding a latent vector such that if it passes through the generator, it generates an image similar to the query image. Finally, the predicted class is selected based on the similarity loss.  We hypothesize that the high-to-low-dimensional mapping in feed-forward classifiers allows adversaries to find small perturbations in the input space that lead to large changes in the output space. On the other hand, a cGAN has a low-dimensional input space. Since the range of images that can be generated by the model for a given class is limited to its learned manifold, the "inversion" process cannot generate images that are arbitrarily close to adversarial examples leading to a robust model by construction. While the method is related to Defense-GAN \cite{samangouei2018defense}, which cleans adversarial examples by projecting to the data manifold prior to passing them to a classifier, the use of the cGAN and inversion in our model instead of the feed-forward classifier is a critical difference. We show that our method does not obfuscate gradients and demonstrate that our model is extremely robust against black-box attacks, and has improved robustness against white-box attacks compared to naturally trained, feed-forward classifiers. 

\section{Related Work}
Adversarial attacks were first noticed by Szegedy et al. \cite{szegedy2013intriguing}. They used the L-BFGS method to find a perturbed image similar to the target image under L2 distance. Later, Goodfellow et al.  \cite{goodfellow2014explaining} introduced the fast gradient sign method (FGSM) as a fast way to generate adversarial examples. This method uses only one step of back-propagation. It adds the sign of gradients to the inputs to change the image to fool the classifier model. PGD (projected gradient descent) \cite{madry2017towards} and BIM (basic iterative method) \cite{kurakin2016adversarial} were introduced as an extension of the FGSM attack. Since they are the multistep variants of FGSM, they are more powerful attacks.
Moosavi-Dezfooli et al. \cite{moosavi2016deepfool} introduced Deepfool for generating adversarial attacks. They generated attacks by adding perturbations to the input to move it beyond the decision boundary.

Carlini and Wagner \cite{carlini2017towards} introduced a powerful type of attack named C\&W attack. They solved an optimization problem to find the minimum change to make to an image to change its predicted label. They showed their attack is successful on defensive distillation \cite{papernot2016distillation}.  
Later, attacks were introduced by Athalye et al. \cite{athalye2018obfuscated} that were designed to defeat defenses that work by masking/obfuscating gradients. Backward pass differentiable approximation (BPDA) is one of the methods that works by approximating the gradients in the backward pass.
Papernot et al. \cite{papernot2017practical} introduced a new way to generate adversarial attacks where the attacker has no access to details of the target model (black-box attacks). The attacker can only feed in input and get the output predicted by the target model. This method works based on the transferability of adversarial attacks. The attacker trains its classifier, called the substitute model, and creates attacks for the substitute model \cite{xu2020adversarial}. \\
To protect DNNs against adversarial examples, researchers have developed different defense techniques, including, 
adversarial training \cite{szegedy2013intriguing, goodfellow2014explaining}. This defense strategy feeds adversarial examples with their correct labels during the training process. This process has been shown to increase the robustness of the model. However, it will not be as effective if it is used with incorrect knowledge of the attacker or a different attack type. 
In other words, this defense mechanism requires previous knowledge about the attacks. Unlike adversarial training,  we propose a classification method that does not make any assumption about the attack type or strength.

Defensive distillation \cite{papernot2016distillation} is another defense method that is based on the distillation idea introduced by Hinton et al. \cite{hinton2015distilling}. They used the extracted information during distillation to reduce the amplitude of the gradients used to craft adversarial examples. Networks trained with the distillation method are less sensitive to adversarial examples. However, this method was broken by C\&W attack. 

Guo et al. \cite{guo2017countering} introduced a defense technique by applying different input transformations to the image to increase the robustness against adversarial attacks. This method was not effective against white-box attacks, and it was bypassed by Athalye et al. \cite{athalye2018obfuscated}. They were successful in reducing the accuracy to 0.

Defense methods such as Defense-GAN \cite{samangouei2018defense} and PixelDefend \cite{song2017pixeldefend} were developed by masking gradients. Defense-GAN was introduced by Samangouei et al. \cite{samangouei2018defense} to purify adversarial images before feeding them to the classifier. The authors used a pre-trained generator on unperturbed images to diminish the adversarial perturbation by projecting images onto the range of the generative model. PixelDefend is similar to Defense-GAN. Instead of a GAN model, it uses a PixelCNN to project adversarial examples back to the data manifold before feeding them to the classifier. 
These two methods appear to be robust against adversarial examples, but they are not robust against attacks that do not use gradients or are successful in approximating the gradients. As an example, BPDA was used to overcome both of these defense techniques. Our method does not obfuscate gradients and is extremely robust against non-gradient-based black-box attacks, and has improved robustness against white-box attacks compared to naturally trained, feed-forward classifiers.

Huang et al. \cite{huang2019defense} presented a reconstruction network called AE-GAN$_{+rs}$ to improve the computational cost of Defense-GAN. They trained an AE-GAN by optimizing adversarial loss in addition to the reconstruction loss. 
Instead of randomly initializing the latent space at inference time, they used the encoder part of autoencoder for latent space initialization to purify images.  This method results in reducing the computational time of Defense-GAN. Even though the results of AE-GAN are comparable with Defense-GAN, this method is also based on gradient masking.

Rezaeifar et al. \cite{rezaeifar2018classification} introduced a  classification method based on reconstruction loss. For each class, they trained a separate variational autoencoder (VAE) and used the binary cross-entropy loss of the input and the reconstructed output to classify (binary) images at inference time. They also performed outlier detection by thresholding based on Kullback-Leibler (KL) divergence. Due to training a separate VAE for each class, this method does not scale well for large-scale datasets. The critical difference of our approach lies in not using reconstruction error but rather the distance between the query and the generated image using optimization.

\section{Method}

\subsection{Inverting a Conditional Generator}
\begin{figure}[!h]
  \centering
  \includegraphics[width=3.5in]{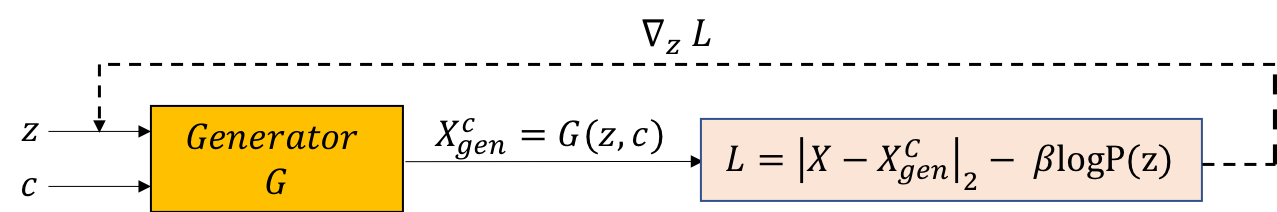}
  \caption{Inverting a conditional generator G using the gradient descent algorithm.}
  \label{inversion_fig}
\end{figure}

\begin{algorithm}[H]
\caption{Inverting generator $G$}
\begin{algorithmic}
\REQUIRE Pretrained conditional generator $G$, target image $X$, input class $c$. 
\STATE Initialize latent space $z^* \sim P(z)$. 
\WHILE {Not converged }
    \STATE 1. $L \leftarrow |X - G(z^*, c)|_2 - \beta logP(z^*)$
    \STATE 2. $z^*  \leftarrow z^* - \alpha \nabla_{z} L$
\ENDWHILE

\STATE Return $z^*$
\end{algorithmic}
\label{algorithm_invert}
\end{algorithm}
Our method involves inverting a conditional generator. We start from a conditional generator trained on unperturbed images, $G(z,c)$, where $z \in Z$ and c is the class label. For a query image $x$ to be classified, for each possible class $c$, we aim to find a $z$ that produces an image $x^c_{gen}$ similar to $x$ where similarity is defined as $L_2$ distance in pixel space. Fig.~\ref{inversion_fig} provides a general overview of the inversion process. We follow a procedure similar to the one provided by Creswell and Bharath \cite{creswell2018inverting} for inverting a generator. The difference is that we invert a conditional generator rather than a generator. The optimal $z^*$ can be found by  

\begin{equation}
L_{gen}(x, G(z,c)) = |G(z,c) - x|_2
\end{equation}

\begin{equation}
z^{\ast} = min_{z} L_{gen}(x, G(z,c))
\end{equation}

Since equation (1) is differentiable, gradient descent-based algorithms can be used to minimize it. Algorithm~\ref{algorithm_invert} provides the details of the inversion process. 

In addition to minimizing the generative loss in equation (1), we also maximize the likelihood of $z$ under a prior distribution that was used to sample $z$ during the training of the cGAN \cite{creswell2018inverting}.  For the $d$-dimensional latent vector $z$ with i.i.d elements drawn from a Gaussian distribution, the log-likelihood of $z$ is calculated as 
\begin{equation}
logP(z) = logP(z^1, ..., z^d) = 1/d \sum_{i=1}^{i=d}logP(z^i).
\end{equation} 
Therefore, the total loss becomes
\begin{equation}
L(z) = L_{gen}(x, G(z,c)) -\beta logP(z),  
\end{equation}
where $\beta$ is a weighting parameter. Using the total loss in (4), $z$ is  obtained by optimizing   
\begin{equation}
z^{\ast} = min_{z} L(z). 
\end{equation}
In this paper, the number of gradient descent iterations is specified by $T$. Also, since $z$ is initialized randomly, we perform the inversion process for $N$ different times to obtain the best generated image.

\subsection{Classification By Inversion} 
\label{sec:classification}
\begin{figure}[!h]
  \centering
  \includegraphics[width=3.55in]{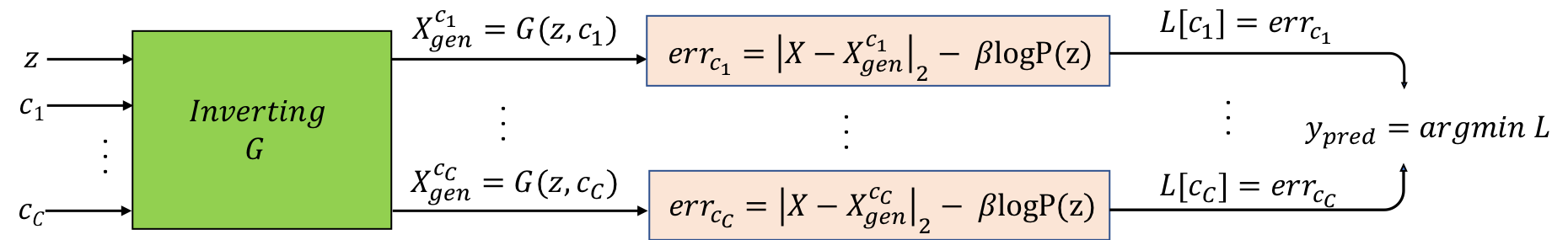}
  \caption{Classifying an image $x$ by inverting a conditional generator G. }
  \label{general_algo}
\end{figure}

In this section, we explain how we utilize an inverted conditional generator to classify an input image. Fig.~\ref{general_algo} illustrates the general process of our algorithm.  We start by feeding the generator a randomly initialized $z$ and with a  class label $c_i$, $i \in \{ 1, ..., C\}$, where $C$ is the total number of classes. For class, we perform the inversion using equation (5). For each of generated $x^{c_i}_{gen}$, we compute the total loss in equation (4). Finally, we select a class label corresponding to the lowest loss value: 
\begin{equation}
y_{pred} = argmin_{c \in C} (|G(z,c) - x|_2 -\beta logP(z))
\end{equation}

\begin{figure*}[!h]
  \centering
  \includegraphics[width=115mm, height=45mm]{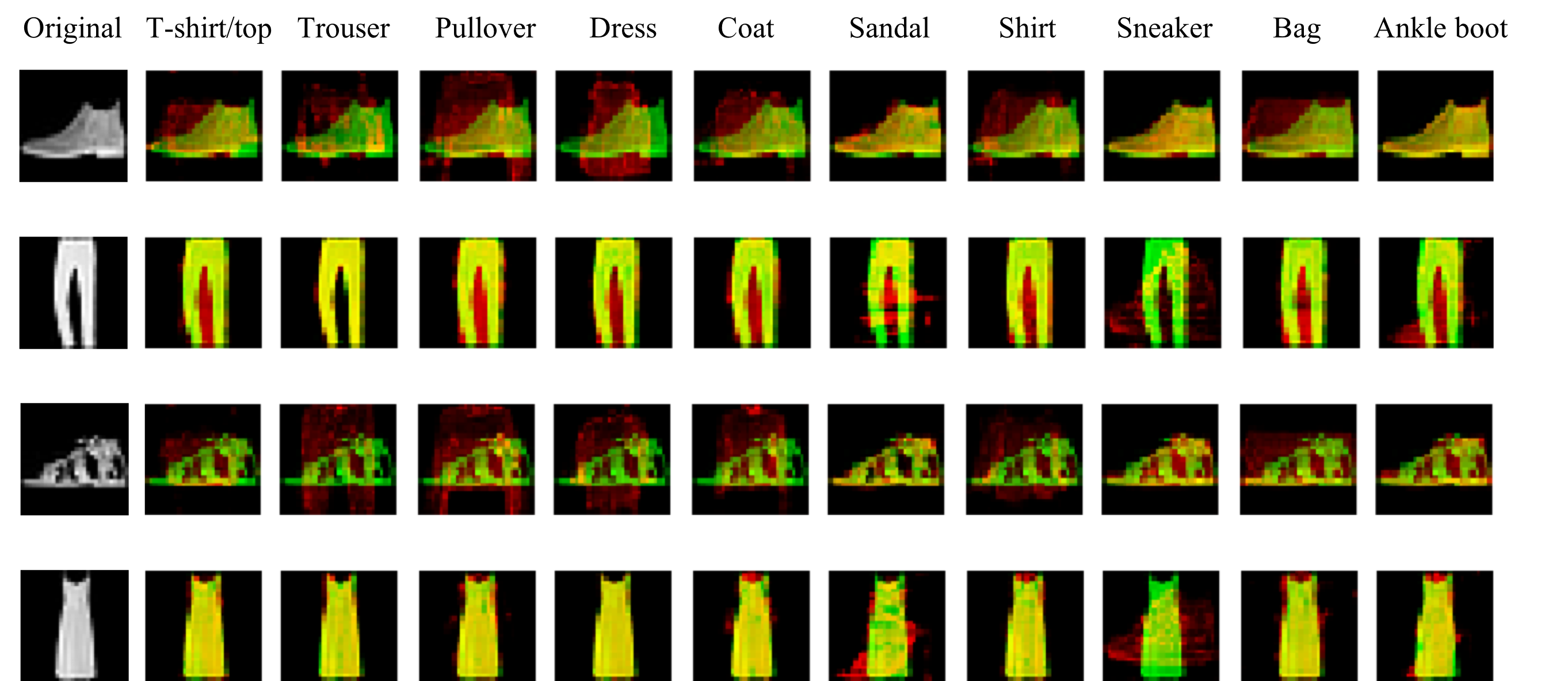}
  \caption{Examples of generated images from FMNIST testing dataset. The first column shows the original images. The second to the last column shows the overlap of original images with the generated images by reversing the conditional generator.  As shown, the target images have the most overlap with generated images with correct class. True labels from top to bottom: ankle boot, trouser, sandal, dress. }
  \label{examples_class}
\end{figure*}

\begin{figure*}[!h]
  \centering
  \includegraphics[width=115mm, height=45mm]{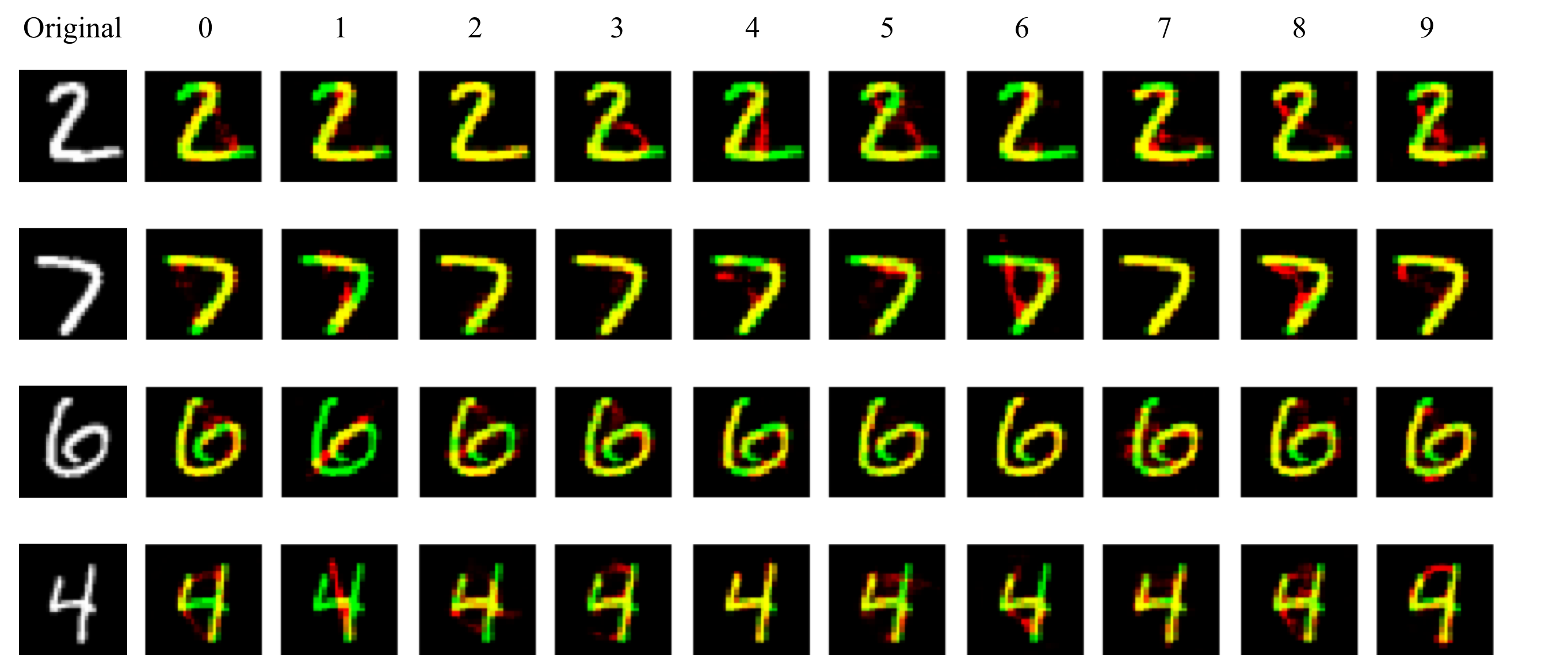}
  \caption{Examples of generated images from MNIST testing dataset. The first column shows the original images. The second to the last column shows the overlap of original images with the generated images by reversing the conditional generator.  The target images have the most overlap with generated images with correct class.}
  \label{examples_mnist_class}
\end{figure*}

We provide the details of our classification method in Algorithm~\ref{algorithm_genral}. In Figures ~\ref{examples_class} and ~\ref{examples_mnist_class}, we visualize examples of generated images for classification for the MNIST and FMNIST datasets. Each row corresponds to an example followed by generated images by changing the class in the inversion process. As shown, the original images have the most overlap with generated images with correct class.

It is expected that a gradient-based adversary would use gradient ascent on $|G(z,c) - x|_2$ with respect to the query image $x$ for the correct class $c$ and gradient descent on the same for either all incorrect classes or a specific class for a targeted attack. The gradient of $|G(z,c) - x|_2$ with respect to $x$ is found as 
\begin{equation}
\label{eqn:obfuscate}
    \frac{\partial |G(z,c) - x|_2}{\partial x}=
    (G(z,c) - x)
    \left(
    \frac{\partial G(z,c)}{\partial x} - 1
    \right),
\end{equation}
which is directly proportional to $(G(z,c) - x)$ and is not obfuscated. This is due to the presence of $x$ in the decision function which leads to the a usable gradient even if $\partial G / \partial x$ is unintentionally obfuscated. This is different from Defense-GAN which passes the output of an unconditional generator $G(z)$ to a classifier, i.e. $f(G(z))$. Since a set of images $x$ will have the same closest point $G(z)$ on the generator manifold, it is expected that the gradient of $G(z)$ with respect to $x$ will be very small almost everywhere. Consequently, the gradient of $f$ with respect to $x$ is obfuscated due to multiplication with the gradient of $G(z)$ in the chain rule.

\begin{algorithm}[H]
\caption{Classification by inversion}
\begin{algorithmic}
\REQUIRE Pretrained conditional generator $G$, target image $X$, number of classes $C$.
\STATE Initialize empty error set $L \leftarrow  \{\} $.
\FOR {each c in C}
\STATE $err_{N} \leftarrow  \{\} $
\FOR {n in N}
    \STATE  $z^* \leftarrow $ invert $G$ given $X$ and $c$
    \STATE  $X^c_{n} \leftarrow G(z^* , c) $
    \STATE  $err_{N} \leftarrow err_{N}  \cup |X - X^c_{n}|_2 - \beta logP(z^*)$
\ENDFOR
    \STATE  $err_c \leftarrow min$  $err_{N}$
    \STATE  $L \leftarrow L \cup err_c$
\ENDFOR

$y_{pred} \leftarrow argmin$  $L $
\STATE Return $y_{pred}$
\end{algorithmic}
\label{algorithm_genral}
\end{algorithm}

\subsection{Architecture of cGAN}
Conditional generative adversarial networks (cGAN) were introduced by Mirza and Osindero \cite{mirza2014conditional} as an extended version of generative adversarial networks (GAN) \cite{goodfellow2014generative}. In a cGAN, both generator $G$ and discriminator $D$ are conditioned on extra information $y$ (e.g., class labels). Training $G$ and $D$ is done by optimizing the following min-max value function: 
\begin{equation}
\begin{aligned}
&min_G max_D V(D,G) = E_{x \sim p_{data}(x)} [log D(x|y)] + \\
&E_{z \sim p_{z}(z)} [log(1 - D(G(z|y)))].
\end{aligned}
\end{equation}

Due to the difficult training process of GANs, Arjovsky et al. \cite{arjovsky2017wasserstein} introduced Wasserstien GAN (WGAN) with an stable training process, where the following function is optimized:
\begin{equation}
\begin{aligned}
&min_G max_D V(D,G) = E_{x \sim p_{data}(x)} [D_w(x|y)] - \\ 
&E_{z \sim p_{z}(z)} [D_w(G(z|y))].
\end{aligned}
\end{equation}

In this paper, we use a WGAN with gradient penalty (GP) introduced in \cite{gulrajani2017improved}. We provide the detailed architecture of the cGAN in Appendix A.

\section{Experiments}
\subsection{Setting and Datasets}
In order to show the robustness of our method, we perform experiments on MNIST \cite{lecun1998gradient} and FMNIST \cite{xiao2017fashion} datasets. 
These datasets contain 60000 training images and 10000 testing images. 
We compare our method to Defense-GAN against white-box and black-box attacks. 
For black-box attacks we consider FGSM \cite{goodfellow2014explaining} and PGD \cite{madry2017towards} attacks. For white-box attacks, in addition to the FGSM and PGD attacks , we also perform BPDA \cite{athalye2018obfuscated} attack on Defense-GAN and report the results.
We train the cGAN for our model using WGAN training formulation \cite{gulrajani2017improved}.
For black box attacks, we divide the testing dataset into a small holdout set of 160 images for training the substitute model \cite{papernot2017practical} and the remaining 9840 images for testing against different adversarial attacks. 
For white-box attacks, we test on all 10000 images.  In addition to the main attack results, we also explore the performance of our model by varying the number of initialization and number of optimization iterations in ablation study section. For all the experiments in this paper, we choose $T=200, N=10$ and $\beta = 0.5$.
All implementations\footnote{Our code will be released under MIT license} are done in Pytorch \cite{paszke2019pytorch} on TITANX GPUs. 
For adversarial attacks implementation, we use the Advertorch library \cite{ding2019advertorch} in Pytorch. 

\subsection{Results of Black-Box Attacks}
In this section, we report the results of FGSM and PGD black-box attacks. 
The black-box attacks are done as described in \cite{papernot2017practical}.
The attacker first trains a substitute model on a small labeled dataset (here 160 images) and continues the training by querying labels from the target model. This process gives the attacker the decision boundary of the target model. After the substitute training is completed, white-box attacks are generated using the substitute model. Since the adversarial attacks are transferable, they will be effective on the target model as well. 
For FGSM attacks, the architectures of substitute models B and E are the same as described in \cite{samangouei2018defense}.

In Tables\footnote{Results of Defense-GAN and Adv. Tr. are taken from \cite{samangouei2018defense}}~\ref{BB-FGSM-MNIST} and ~\ref{BB-FGSM-FMNIST}, we present the classification accuracy under black-box FGSM attack with $\epsilon=0.3$ on MNIST and FMNIST datasets.

The accuracy of our method is better (up to 17.96\%) than Defense-GAN across different substitute models and different datasets. Adversarial training performs well only with the correct knowledge of $\epsilon$, and its accuracy decreases with incorrect $\epsilon$. As shown in Tables~\ref{BB-FGSM-MNIST} and ~\ref{BB-FGSM-FMNIST}, We achieve comparable results with adversarial training with $\epsilon=0.3$ on the MNIST dataset. We also outperform adversarial training by a large margin (up to 11.56\%) for the FMNIST dataset.
Our method generally outperforms both Defense-GAN and adversarial training on black-box attacks because it does not require any previous knowledge on attacks. 
\begin{table} [!h]
  \caption{Black-box PGD attack with  $\epsilon=0.3$, $0.4$ on MNIST}
  \label{BB-PDG-MNIST}
  \centering
  \begin{tabular}{llll}
    \toprule
    Substitute model     & No attack     & PGD, $\epsilon=0.3$ &  PGD, $\epsilon=0.4$\\
    \midrule
    Sub. A & 97.50 & 95.76  & 94.94\\
    \bottomrule
  \end{tabular}
\end{table}

\begin{table*}[!h]
  \caption{Black-box FGSM attack with $\epsilon=0.3$ on MNIST}
  \label{BB-FGSM-MNIST}
  \centering
  \begin{tabular}{llllll}
    \toprule
    Substitute models     & No attack     & Our method &  Defense-GAN  &Adv. Tr., $\epsilon=0.3$ & Adv. Tr., $\epsilon=0.15$  \\
    \midrule
    Sub. B & 97.50 & 96.09 & 93.12 & \textbf{96.54} & 62.23 \\
    Sub. E     & 97.50 & 95.63 & 91.39 & \textbf{96.68} & 93.27 \\

    \bottomrule
  \end{tabular}
\end{table*}

\begin{table*} [!h]
  \caption{Black-box FGSM attack with $\epsilon=0.3$ on FMNIST}
  \label{BB-FGSM-FMNIST}
  \centering
  \begin{tabular}{llllll}
    \toprule
    Substitute model     & No attack     & Our method &  Defense-GAN  & Adv. Tr., $\epsilon=0.3$  & Adv. Tr., $\epsilon=0.15$\\
    \midrule
    Sub. B & 84.48 & \textbf{77.56}  & 58.60 & 73.93 & 66.00 \\
    Sub. E     & 84.48 & \textbf{72.13} & 47.90 & 69.45 & 56.38 \\

    \bottomrule
  \end{tabular}
\end{table*}
To show the robustness of our model against stronger black-box attacks, we also report the accuracy of our model on PGD attacks. The PGD attack is generated with $\epsilon=0.3$ and $\epsilon=0.4$ for 40 iterations with step size $0.01$ on 1500 images of the MNIST dataset.

As we see in Table~\ref{BB-PDG-MNIST}, our method performs well against PGD attack. We noticed that the accuracy drops by only 0.82\% as $\epsilon$ increases to $0.4$. 
These experiments confirm the robustness of our method against black-box attacks by construction where no previous knowledge on attack type nor strength is required.

\subsection{Results of White-Box Attacks}
In this section, we show the accuracy results under white-box FGSM and PGD attack with $\epsilon=0.3$. We maximize a hinge loss function using gradient ascent for white-box attacks: 
\begin{equation}
\mathcal{L}(\mathcal{S}, y) = \frac{1}{C} \sum_{i, i \neq y }^{C}(max(0, 1 - \mathcal{S}(y) +\mathcal{S}(i)),
\end{equation}
where $C$ is number of classes, $y$ is the true label, and $\mathcal{S} = - L(z) $ obtained from equation (4).

For comparison, we also perform BPDA attack for the Defense-GAN method. As Athalye et al.\cite{athalye2018obfuscated} showed, BPDA can reduce the accuracy of Defense-GAN under adversarial attacks by 45\% on the MNIST dataset \cite{athalye2018obfuscated} and 59\% on the FMNIST dataset. These experiments show that although Defense-GAN appears to be robust against white-box attacks, BPDA attack can circumvent the defense and lower the accuracy significantly. The BPDA attack is not applicable for our method because the gradients of our method are already unobfuscated as shown in Section~\ref{sec:classification}. The substitution of the gradient of $G$ with the identity function employed by BPDA in (\ref{eqn:obfuscate}) would result in an unusable $0$ gradient. On the other hand, replacing $\frac{\partial G(z,c)}{\partial x} - 1$ in (\ref{eqn:obfuscate}) with the identity function makes no change to the gradient when the gradient of $G$ is obfuscated, i.e. when $\frac{\partial G(z,c)}{\partial x}\approx 0$. As shown in Tables\footnote{Results of Defense-GAN and Adv. Tr. are taken from \cite{huang2019defense}}~\ref{WB-MNIST} and ~\ref{WB-fMNIST}, our method significantly outperforms Defense-GAN under BPDA attack. Adversarial training outperforms our method for both FGSM and PGD attacks. 

\begin{table} [!h]
  \caption{White-box attacks with $\epsilon=0.3$ on MNIST.}
  \label{WB-MNIST}
  \centering
  \begin{tabular}{lllll}
    \toprule
    Attack type     & No attack     & Our method &  Defense-GAN & Adv. Tr. \\
    \midrule
    FGSM & 97.50 & 86.38 & 98.10 & \textbf{94.90} \\
    PGD  & 97.50 & 85.77 & 98.90 & \textbf{92.00} \\
    BPDA  & - & -& 55.00 & - \\

    \bottomrule
  \end{tabular}
\end{table}


\begin{table} [!h]
  \caption{White-box attacks with $\epsilon=0.3$ on FMNIST.}
  \label{WB-fMNIST}
  \centering
  \begin{tabular}{lllll}
    \toprule
    Attack type     & No attack     & Our method &  Defense-GAN & Adv. Tr. \\
    \midrule
    FGSM & 84.48 & 41.80 & 81.40 &  \textbf{73.90} \\
    PGD  & 84.48 & 36.92 & 85.20 &  \textbf{71.70} \\
    BPDA  & - & - & 26.61 & - \\
    \bottomrule
  \end{tabular}
\end{table}
\subsection{Ablation Study}
\subsubsection{Effect of $LogP$ on Accuracy}
In this section, we study the effect of our loss function components in equation (4). To this end, we perform classification with and without including $logP$ in the inversion and class selection process. 

In Table~\ref{LOGP},  we report the classification accuracy for MNIST and FMNIST datasets with no attack. We observe that using $LogP$ in equation (4) has the expected regularization effect and improves the classification accuracy for both MNIST and FMNIST datasets.

\begin{table} [!h]
  \caption{Effect of $LogP$ on classification accuracy}
  \label{LOGP}
  \centering
  \begin{tabular}{lll}
    \toprule
    Accuracy     & With $LogP$     & Without $LogP$\\
    \midrule
    MNIST & 97.50 & 86.71\\
    FMNIST  & 84.48 & 82.08 \\
    \bottomrule
  \end{tabular}
\end{table}

\subsubsection{Effect of Number of Iterations and Initialization on Accuracy}
\begin{figure*}
  \centering
    \subfloat{\includegraphics[width=2.5in]{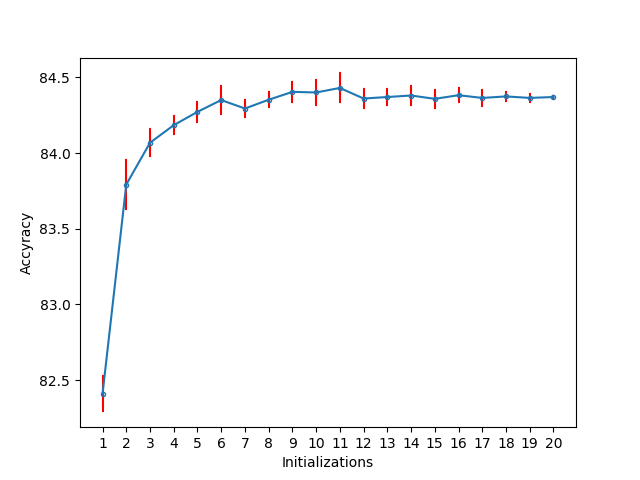}}%
    \subfloat{\includegraphics[width=2.5in]{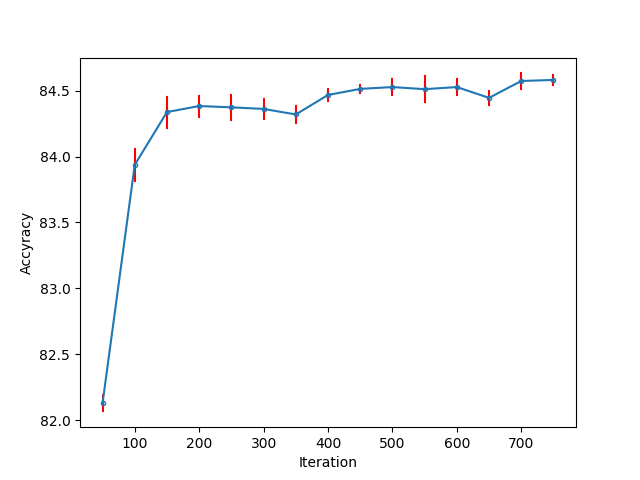}}%
  \caption{Classification accuracy of our model on MNIST dataset with no attack. Left: accuracy vs. number of initialization $N$. Right: accuracy vs. number of steps $T$. The accuracy represents mean$\%$ $\pm$ std. dev. over five times  repetition with different random set of $z$ initializations. } 
  \label{acc_vs_init}
\end{figure*}
We investigate the effect of varying the number of random initializations and the number of steps on the classification accuracy of our model with no attack.
As shown in Fig.~\ref{acc_vs_init}, increasing the number of random initialization improves the classification accuracy. We also observe a similar pattern for the accuracy versus the number of optimization steps. 
\section{Conclusion}
In this paper, we have presented a robust classification method that does not obfuscate gradients and requires no previous knowledge about the attack. We utilized a pretrained cGAN on natural images and inverted the generator to find the class that generates the closest image to the query image. Since generative models are low-to-high-dimensional mappings, we eliminated a potential source of vulnerability against adversarial attacks, which is high-to-low-dimensional mapping of natural feed-forward classifiers. 

Our classification method is related to Defense-GAN, but the difference between our method and Defense-GAN lies in using a cGAN and inverting it instead of using a feed-forward classifier. We showed that our method is highly effective against black-box attacks and increases the robustness against white-box attacks compared to naturally trained, feed-forward classifiers. In future work, adversarial training can be combined with our method to create a defense that is robust to both gradient-based and non-gradient-based attacks. 

The performance of our classification method relies on the expressiveness power of the trained generator. Improving the cGAN is expected to improve the classification accuracy. Since the classification is performed by inverting the generator for each possible class, one of the drawbacks of the proposed method is computational expense as the number of classes increases. 

\section*{Acknowledgment}
This work was sponsored by DARPA TRADES Award HR0011-17-2-0016.

\bibliographystyle{IEEEtran}
\bibliography{Ref}

\begin{thebibliography}{10}
\providecommand{\url}[1]{#1}
\csname url@samestyle\endcsname
\providecommand{\newblock}{\relax}
\providecommand{\bibinfo}[2]{#2}
\providecommand{\BIBentrySTDinterwordspacing}{\spaceskip=0pt\relax}
\providecommand{\BIBentryALTinterwordstretchfactor}{4}
\providecommand{\BIBentryALTinterwordspacing}{\spaceskip=\fontdimen2\font plus
\BIBentryALTinterwordstretchfactor\fontdimen3\font minus
  \fontdimen4\font\relax}
\providecommand{\BIBforeignlanguage}[2]{{%
\expandafter\ifx\csname l@#1\endcsname\relax
\typeout{** WARNING: IEEEtran.bst: No hyphenation pattern has been}%
\typeout{** loaded for the language `#1'. Using the pattern for}%
\typeout{** the default language instead.}%
\else
\language=\csname l@#1\endcsname
\fi
#2}}
\providecommand{\BIBdecl}{\relax}
\BIBdecl

\bibitem{krizhevsky2012imagenet}
A.~Krizhevsky, I.~Sutskever, and G.~E. Hinton, ``Imagenet classification with
  deep convolutional neural networks,'' \emph{Advances in neural information
  processing systems}, vol.~25, pp. 1097--1105, 2012.

\bibitem{DBLP:journals/corr/SimonyanZ14a}
\BIBentryALTinterwordspacing
K.~Simonyan and A.~Zisserman, ``Very deep convolutional networks for
  large-scale image recognition,'' in \emph{3rd International Conference on
  Learning Representations, {ICLR} 2015, San Diego, CA, USA, May 7-9, 2015,
  Conference Track Proceedings}, Y.~Bengio and Y.~LeCun, Eds., 2015. [Online].
  Available: \url{http://arxiv.org/abs/1409.1556}
\BIBentrySTDinterwordspacing

\bibitem{DBLP:journals/corr/SzegedyLJSRAEVR14}
\BIBentryALTinterwordspacing
C.~Szegedy, W.~Liu, Y.~Jia, P.~Sermanet, S.~E. Reed, D.~Anguelov, D.~Erhan,
  V.~Vanhoucke, and A.~Rabinovich, ``Going deeper with convolutions,''
  \emph{CoRR}, vol. abs/1409.4842, 2014. [Online]. Available:
  \url{http://arxiv.org/abs/1409.4842}
\BIBentrySTDinterwordspacing

\bibitem{lecun2015deep}
Y.~LeCun, Y.~Bengio, and G.~Hinton, ``Deep learning,'' \emph{nature}, vol. 521,
  no. 7553, pp. 436--444, 2015.

\bibitem{Ren2015}
S.~Ren, K.~He, R.~Girshick, and J.~Sun, ``Faster r-cnn: Towards real-time
  object detection with region proposal networks,'' in \emph{Advances in Neural
  Information Processing Systems (NeurIPS)}, 2015.

\bibitem{Redmon2016}
J.~Redmon, S.~Divvala, R.~Girshick, and A.~Farhadi, ``You only look once:
  Unified, real-time object detection,'' in \emph{CVPR}, 2016.

\bibitem{he2016deep}
K.~He, X.~Zhang, S.~Ren, and J.~Sun, ``Deep residual learning for image
  recognition,'' in \emph{Proceedings of the IEEE conference on computer vision
  and pattern recognition}, 2016, pp. 770--778.

\bibitem{szegedy2013intriguing}
C.~Szegedy, W.~Zaremba, I.~Sutskever, J.~Bruna, D.~Erhan, I.~Goodfellow, and
  R.~Fergus, ``Intriguing properties of neural networks,'' \emph{arXiv preprint
  arXiv:1312.6199}, 2013.

\bibitem{goodfellow2014explaining}
I.~J. Goodfellow, J.~Shlens, and C.~Szegedy, ``Explaining and harnessing
  adversarial examples,'' \emph{arXiv preprint arXiv:1412.6572}, 2014.

\bibitem{athalye2018obfuscated}
A.~Athalye, N.~Carlini, and D.~Wagner, ``Obfuscated gradients give a false
  sense of security: Circumventing defenses to adversarial examples,'' in
  \emph{International Conference on Machine Learning}.\hskip 1em plus 0.5em
  minus 0.4em\relax PMLR, 2018, pp. 274--283.

\bibitem{papernot2017practical}
N.~Papernot, P.~McDaniel, I.~Goodfellow, S.~Jha, Z.~B. Celik, and A.~Swami,
  ``Practical black-box attacks against machine learning,'' in
  \emph{Proceedings of the 2017 ACM on Asia conference on computer and
  communications security}, 2017, pp. 506--519.

\bibitem{guo2017countering}
C.~Guo, M.~Rana, M.~Cisse, and L.~Van Der~Maaten, ``Countering adversarial
  images using input transformations,'' \emph{arXiv preprint arXiv:1711.00117},
  2017.

\bibitem{samangouei2018defense}
P.~Samangouei, M.~Kabkab, and R.~Chellappa, ``Defense-gan: Protecting
  classifiers against adversarial attacks using generative models,''
  \emph{arXiv preprint arXiv:1805.06605}, 2018.

\bibitem{song2017pixeldefend}
Y.~Song, T.~Kim, S.~Nowozin, S.~Ermon, and N.~Kushman, ``Pixeldefend:
  Leveraging generative models to understand and defend against adversarial
  examples,'' \emph{arXiv preprint arXiv:1710.10766}, 2017.

\bibitem{xu2020adversarial}
H.~Xu, Y.~Ma, H.-C. Liu, D.~Deb, H.~Liu, J.-L. Tang, and A.~K. Jain,
  ``Adversarial attacks and defenses in images, graphs and text: A review,''
  \emph{International Journal of Automation and Computing}, vol.~17, no.~2, pp.
  151--178, 2020.

\bibitem{ma2018characterizing}
X.~Ma, B.~Li, Y.~Wang, S.~M. Erfani, S.~Wijewickrema, G.~Schoenebeck, D.~Song,
  M.~E. Houle, and J.~Bailey, ``Characterizing adversarial subspaces using
  local intrinsic dimensionality,'' \emph{arXiv preprint arXiv:1801.02613},
  2018.

\bibitem{dhillon2018stochastic}
G.~S. Dhillon, K.~Azizzadenesheli, Z.~C. Lipton, J.~Bernstein, J.~Kossaifi,
  A.~Khanna, and A.~Anandkumar, ``Stochastic activation pruning for robust
  adversarial defense,'' \emph{arXiv preprint arXiv:1803.01442}, 2018.

\bibitem{xie2017mitigating}
C.~Xie, J.~Wang, Z.~Zhang, Z.~Ren, and A.~Yuille, ``Mitigating adversarial
  effects through randomization,'' \emph{arXiv preprint arXiv:1711.01991},
  2017.

\bibitem{buckman2018thermometer}
J.~Buckman, A.~Roy, C.~Raffel, and I.~Goodfellow, ``Thermometer encoding: One
  hot way to resist adversarial examples,'' in \emph{International Conference
  on Learning Representations}, 2018.

\bibitem{mirza2014conditional}
M.~Mirza and S.~Osindero, ``Conditional generative adversarial nets,''
  \emph{arXiv preprint arXiv:1411.1784}, 2014.

\bibitem{creswell2018inverting}
A.~Creswell and A.~A. Bharath, ``Inverting the generator of a generative
  adversarial network,'' \emph{IEEE transactions on neural networks and
  learning systems}, vol.~30, no.~7, pp. 1967--1974, 2018.

\bibitem{madry2017towards}
A.~Madry, A.~Makelov, L.~Schmidt, D.~Tsipras, and A.~Vladu, ``Towards deep
  learning models resistant to adversarial attacks,'' \emph{arXiv preprint
  arXiv:1706.06083}, 2017.

\bibitem{kurakin2016adversarial}
A.~Kurakin, I.~Goodfellow, S.~Bengio \emph{et~al.}, ``Adversarial examples in
  the physical world,'' 2016.

\bibitem{moosavi2016deepfool}
S.-M. Moosavi-Dezfooli, A.~Fawzi, and P.~Frossard, ``Deepfool: a simple and
  accurate method to fool deep neural networks,'' in \emph{Proceedings of the
  IEEE conference on computer vision and pattern recognition}, 2016, pp.
  2574--2582.

\bibitem{carlini2017towards}
N.~Carlini and D.~Wagner, ``Towards evaluating the robustness of neural
  networks,'' in \emph{2017 ieee symposium on security and privacy (sp)}.\hskip
  1em plus 0.5em minus 0.4em\relax IEEE, 2017, pp. 39--57.

\bibitem{papernot2016distillation}
N.~Papernot, P.~McDaniel, X.~Wu, S.~Jha, and A.~Swami, ``Distillation as a
  defense to adversarial perturbations against deep neural networks,'' in
  \emph{2016 IEEE symposium on security and privacy (SP)}.\hskip 1em plus 0.5em
  minus 0.4em\relax IEEE, 2016, pp. 582--597.

\bibitem{hinton2015distilling}
G.~Hinton, O.~Vinyals, and J.~Dean, ``Distilling the knowledge in a neural
  network,'' \emph{arXiv preprint arXiv:1503.02531}, 2015.

\bibitem{huang2019defense}
W.~Huang, S.~Tu, and L.~Xu, ``Defense against adversarial examples by
  encoder-assisted search in the latent coding space,'' 2019.

\bibitem{rezaeifar2018classification}
S.~Rezaeifar, O.~Taran, and S.~Voloshynovskiy, ``Classification by
  re-generation: towards classification based on variational inference,'' in
  \emph{2018 26th European Signal Processing Conference (EUSIPCO)}.\hskip 1em
  plus 0.5em minus 0.4em\relax IEEE, 2018, pp. 2005--2009.

\bibitem{goodfellow2014generative}
I.~J. Goodfellow, J.~Pouget-Abadie, M.~Mirza, B.~Xu, D.~Warde-Farley, S.~Ozair,
  A.~Courville, and Y.~Bengio, ``Generative adversarial networks,'' \emph{arXiv
  preprint arXiv:1406.2661}, 2014.

\bibitem{arjovsky2017wasserstein}
M.~Arjovsky, S.~Chintala, and L.~Bottou, ``Wasserstein generative adversarial
  networks,'' in \emph{International conference on machine learning}.\hskip 1em
  plus 0.5em minus 0.4em\relax PMLR, 2017, pp. 214--223.

\bibitem{gulrajani2017improved}
I.~Gulrajani, F.~Ahmed, M.~Arjovsky, V.~Dumoulin, and A.~Courville, ``Improved
  training of wasserstein gans,'' \emph{arXiv preprint arXiv:1704.00028}, 2017.

\bibitem{lecun1998gradient}
Y.~LeCun, L.~Bottou, Y.~Bengio, and P.~Haffner, ``Gradient-based learning
  applied to document recognition,'' \emph{Proceedings of the IEEE}, vol.~86,
  no.~11, pp. 2278--2324, 1998.

\bibitem{xiao2017fashion}
H.~Xiao, K.~Rasul, and R.~Vollgraf, ``Fashion-mnist: a novel image dataset for
  benchmarking machine learning algorithms,'' \emph{arXiv preprint
  arXiv:1708.07747}, 2017.

\bibitem{paszke2019pytorch}
A.~Paszke, S.~Gross, F.~Massa, A.~Lerer, J.~Bradbury, G.~Chanan, T.~Killeen,
  Z.~Lin, N.~Gimelshein, L.~Antiga \emph{et~al.}, ``Pytorch: An imperative
  style, high-performance deep learning library,'' \emph{arXiv preprint
  arXiv:1912.01703}, 2019.

\bibitem{ding2019advertorch}
G.~W. Ding, L.~Wang, and X.~Jin, ``{AdverTorch} v0.1: An adversarial robustness
  toolbox based on pytorch,'' \emph{arXiv preprint arXiv:1902.07623}, 2019.

\end{thebibliography}

\end{document}